\documentclass[conference]{IEEEtran}
\IEEEoverridecommandlockouts

\usepackage{amsmath,amssymb,amsfonts}
\usepackage{algorithmic}
\usepackage{graphicx}
\usepackage{textcomp}
\usepackage{xcolor}
\def\BibTeX{{\rm B\kern-.05em{\sc i\kern-.025em b}\kern-.08em
    T\kern-.1667em\lower.7ex\hbox{E}\kern-.125emX}}
\begin{document}

\title{Vision-RADAR fusion for Robotics BEV Detections: A Survey\\
\thanks{This review was done while working at Perception team at Motional.}
}

\author{\IEEEauthorblockN{Apoorv Singh}
\IEEEauthorblockA{\textit{Perception Team} \\
\textit{Motional}\\
Pittsburgh, USA \\
apoorv.singh@motional.com}
}

\maketitle

\begin{abstract}
Due to the trending need of building autonomous robotic perception system, sensor fusion has attracted a lot of attention amongst researchers and engineers to make best use of cross-modality information. However, in order to build a robotic platform at scale we need to emphasize on autonomous robot platform bring-up cost as well. Cameras and radars, which inherently includes complementary perception information, has potential for developing autonomous robotic platform at scale. However, there is a limited work around radar fused with Vision, compared to LiDAR fused with vision work. In this paper, we tackle this gap with a survey on Vision-Radar fusion approaches for a BEV object detection system. First we go through the background information viz., object detection tasks, choice of sensors, sensor setup, benchmark datasets and evaluation metrics for a robotic perception system. Later, we cover per-modality (Camera and RADAR) data representation, then we go into detail about sensor fusion techniques based on sub-groups viz., early-fusion, deep-fusion, and late-fusion to easily understand the pros and cons of each method. Finally, we propose possible future trends for vision-radar fusion to enlighten future research.  
Regularly updated summary can be found at: \emph{https://github.com/ApoorvRoboticist/Vision-RADAR-Fusion-BEV-Survey}
\end{abstract}


\section{Introduction}
SAE (Society of Automotive Engineers) has divided the roles of a human driver and driving automation capabilities by the levels of automation viz., Level 0: No driving automation; Level 1: Driver assistance; Level 2: Partial driving automation; Level 3: Conditional driving automation; Level 4: High driving automation; Level 5: Full driving automation. 3D object detection is an essential task for autonomous driving for Level 2 on-wards. However, in order to make these robotic platform at large-scale we need to emphasize on affordable active-saftey hard-wares. Camera and radar perception sensors setup is a low-cost, high-reliability, and low-maintenance one. It can provide rich semantic information with cameras; and long-range detections that are robust to lighting/ weather conditions with radars. LiDAR is a popular choice of sensor for Level 4+ cars however, cameras and radars have dominated L2-L3 levels cars that have been in production over a decade. Recently a lot of interesting work has been explored to utilize camera-radar combination in higher level of automation like  \cite{radar_camera_early}, \cite{grif_net}, and \cite{centerfusion}. 

\begin{figure}[ht]
\vskip 0.2in
\begin{center}
\centerline{\includegraphics[width=\columnwidth]{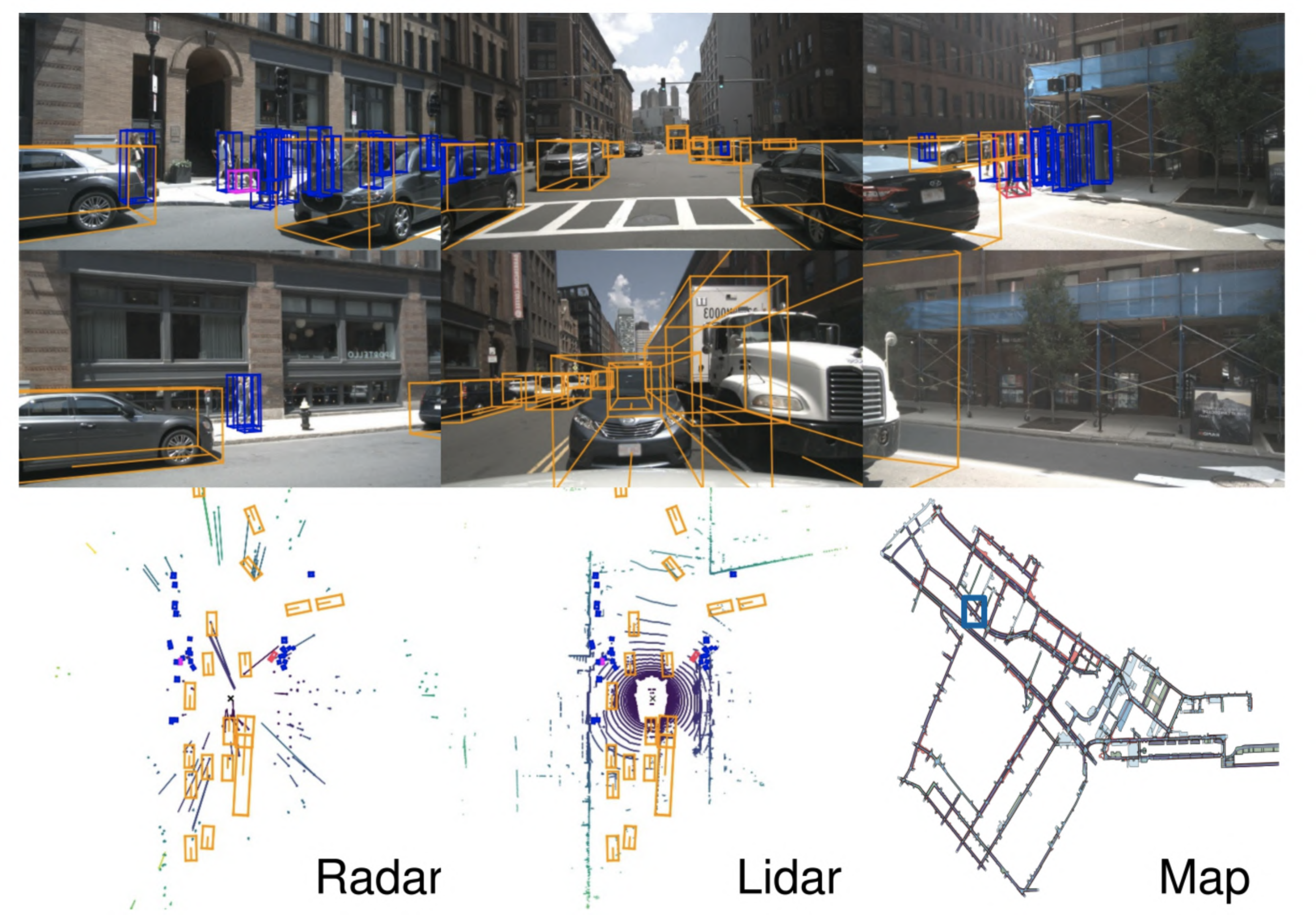}}
\caption{BEV Perception with camera, radar, HD-map and lidar data. Snapshot taken from Multi-modal dataset, nuScenes \cite{nuscenes}.}
\label{nuscenes_data}
\end{center}
\vskip -0.2in
\end{figure}

\begin{figure}[ht]
\vskip 0.2in
\begin{center}
\centerline{\includegraphics[width=\columnwidth]{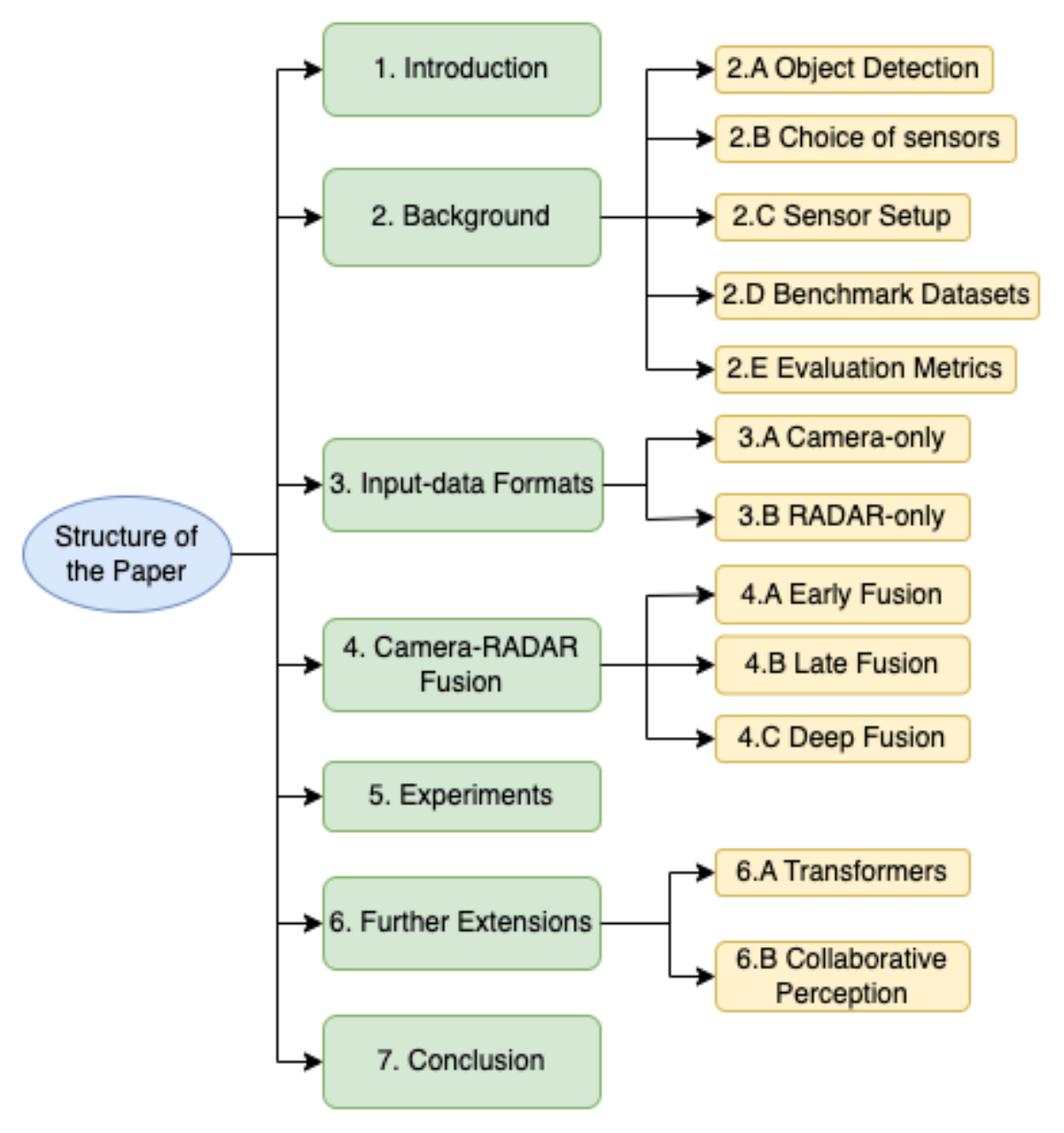}}
\caption{Structure of this Survey Paper.}
\label{sensor-data}
\end{center}
\vskip -0.2in
\end{figure}

Reference \cite{craft} has shown camera and radar's characteristics and how they complement each other. Cameras are not very good in generalizing in BEV predictions, as the input they receive is constrained by 2D pixels. However, they include very rich semantic and boundary information. Radar's data already includes 3D as well as velocity data in the input point-cloud. However, it lacks dense semantic information. For these reasons, camera-radar sensor combination can work very well together, however data received by these sensors need to be mapped to a single coordinate frame. Input data received by them can be visualized in as Fig. \ref{nuscenes_data}

Previous work \cite{multimodal_fusion_22} has only considered vision and lidar aspects. \cite{radar_1} and \cite{radar_2} have covered vision and radar, but they don't dive deep enough on modern deep learning based techniques which are trending in literature these days. With this paper, we plan to target this gap by covering the basics behind BEV detection and senor modalities and then diving deeper into  modern vision-radar fusion techniques giving more focus on the trending transformer based approaches. 

As shown in Fig. \ref{sensor-data}, rest of the paper is organized as follows: We first look at the background information required to understand robotic BEV perception viz., information about Object detection tasks, Choice of sensors, Benchmark datasets, Evaluation metrics etc in section \ref{background_section}. Then, we introduce input-data formats for cameras and radars in section \ref{input_data_section}. In section \ref{cam_radar_section}, we will go through detailed analysis of techniques involved with camera-radar fusion methods. We will also sub-group them so that readers can follow through easily. Later, in section \ref{experiments_section} we will show how discussed methods evaluate on the camera-radar benchmark,  nuScenes \cite{nuscenes}. Then in section \ref{further_extenstion_section} we will go through possible extensions looking at the current research trend that may enlighten future research. Finally in section \ref{conclusion_section} we will conclude our findings.

\section{Background}
\label{background_section}

\subsection{Object Detection Task}
3D Object Detection is an essential task for robotic/ autonomous driving platform. Object detection is a combination of two fundamental computer-vision problems viz., classification and localization. The goal of object detection is to detect all instances of the predefined classes and provide its localization in the image/ BEV space with axis-aligned boxes. It is generally seen as a supervised learning problem which leverages huge amount of labelled images. 
Few of the key challenges in the object detection task include:
\begin{itemize}
\item \emph{Box BEV representation}: Camera images are in perspective-view, however down-the-stream autonomy tasks operate in the Bird's Eye View (BEV). Hence we need a way to transform perspective information to the orthogonal space, BEV. This comes with inherent problem of depth ambiguity, as we are adding a new dimension of depth to this problem. 
\item \emph{Rich Semantic Information}: Sometimes we need to distinguish between very similar looking objects, for example multiple similar looking objects in close vicinity or maybe a pedestrian operating on a skate-board. In later example, pedestrian on skate-board should follow a cyclist's motion-model, but it is a very hard to detect this attribute of a pedestrian. To identify for these fine-grained information we need to embed deep semantics in our model.
\item \emph{Efficiency}: As we are building bigger and deeper networks, we need expensive computation resources to make an deploy-time inference. Edge devices being the common place for deployment platform, it can easily become a bottleneck.
\item \emph{Out of domain objects}: There is a limit to the classes we can train the network with. There will always be some class of object which we may encounter at test time that we haven't seen during training time. We always have lack of some generalization capabilities with detectors. 
\end{itemize}

\subsection{Choice of Sensors}
\begin{figure}[ht]
\vskip 0.2in
\begin{center}
\centerline{\includegraphics[width=\columnwidth]{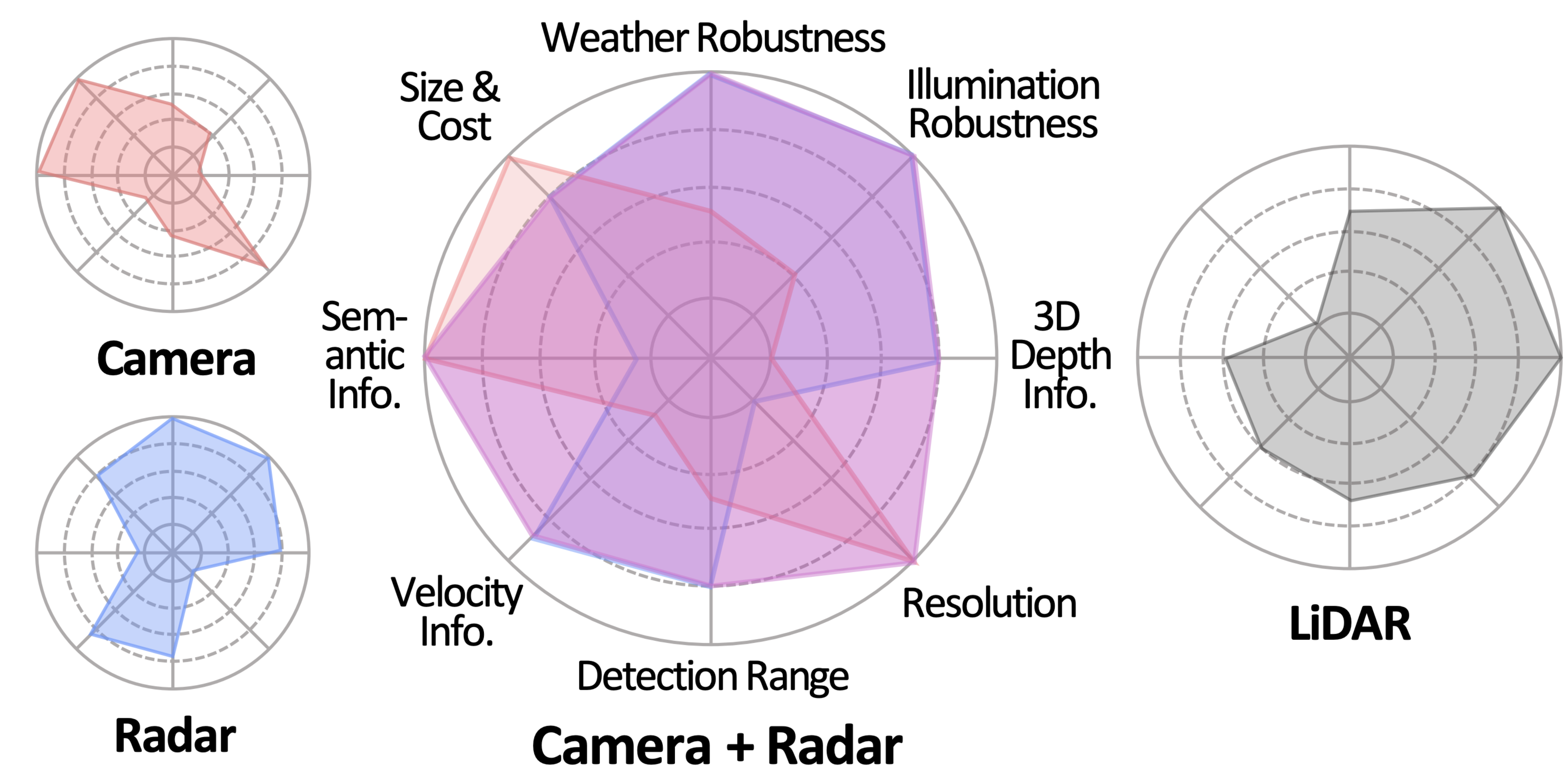}}
\caption{Sensor characteristics of camera, radar, and LiDAR. Camera-radar fusion has high potential considering spatially and contextually complementary properties.}
\label{cam-rada-pros}
\end{center}
\vskip -0.2in
\end{figure}

Cameras and sensors have complementary features, which sets them for robust perception sensor combination. Camera's contribution for detection comes from: Rich semantic information and accurate boundaries. Camera is not very good in fusing temporal data or predicting boxes with accurate depth specially in bad weather conditions. However, radar picks up where camera lags behind. Radars can predict depth and velocity of objects very accurately leveraging doppler effect in their point-cloud. Radar data is very sparse, so it doesn't take too much compute load as well. Radars longer wavelength compared to other laser sensors enables them to be the only perception sensor, for which performance doesn't degrade with adverse weather conditions viz., rain/ snow/ dust etc. These characteristics is very well summarized by \cite{craft} in Fig. \ref{cam-rada-pros}. \\
One of the other less talked issue with radars is its inability to detect velocity components of agents along the radial direction as shown in Fig. \ref{radial_vel_fig}. Another place where radar, and in-fact any laser-based sensor falls behind is detecting black objects/ cars that absorb most of the lasers that falls on them. Camera is the fall-back sensor to rely in these special cases.

\subsection{Sensor Setup}
\begin{figure}[ht]
\vskip 0.2in
\begin{center}
\centerline{\includegraphics[width=\columnwidth]{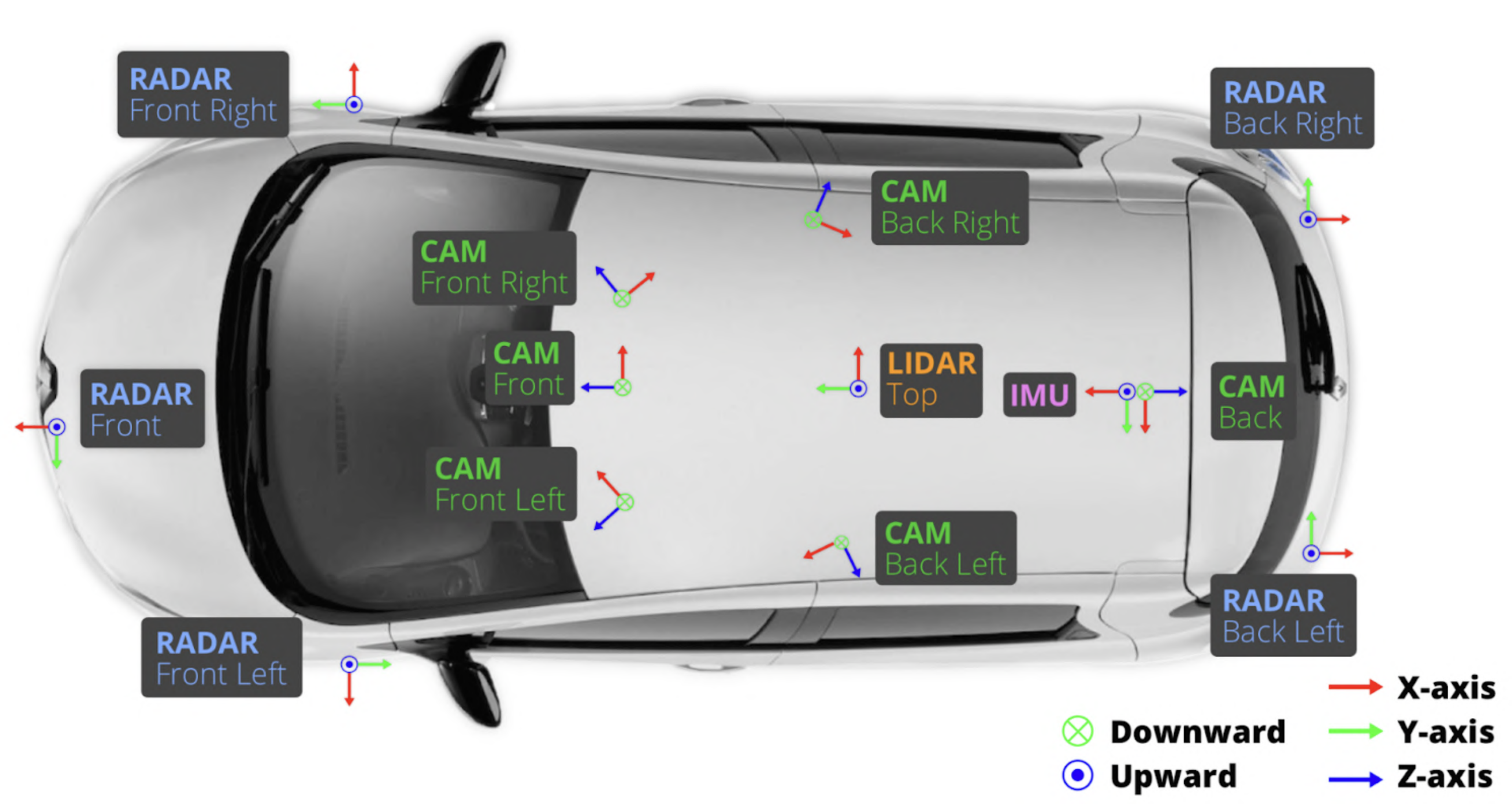}}
\caption{nuScenes \cite{nuscenes} sensor setup.}
\label{nuscenes-setup}
\end{center}
\vskip -0.2in
\end{figure}

There’s a setup of suite of sensors autonomous vehicle (AV), which may vary depending on different autonomous car companies. Typically there are $6-12$ cameras and $3-6$ radars per vehicle. These many sensors are needed to cover the entire surrounding 3D scene. We are limited to use cameras with normal FOV (Field of view) otherwise we may get image distortions that are beyond recovery, like with the Fish-eye cameras (Wide FOV), which are only good for up to few tens of meters. A perception sensor setup in one of the most cited benchmark-dataset, nuScenes \cite{nuscenes} in the AV space can be seen in the Fig. \ref{nuscenes-setup}. 
For the affordability reasons, AV/ mobile-robots industry have always been more invested in radars and cameras in production cars compared to lidars. In this example we see that there are qty. 5 radars, qty. 6 cameras and only qty. 1 lidar. These numbers are representative of other L3+ car companies as well. 

\subsection{Benchmark Datasets}
nuScenes \cite{nuscenes}, KITTI \cite{kitti} and Waymo Open Dataset (WOD) \cite{wod} are the three most commonly used 3D BEV object detection task. Apart from them H3D \cite{h3d}, Lyft L5 \cite{lyftl5_19}, BDD \cite{bdd_18}, STF \cite{stf} and Argoverse \cite{argoverse_19} can also be used for BEV perception tasks. Detailed information on these datasets can be reviewed in Table \ref{dataset_stats}

\begin{table*}[t]
\caption{Information on benchmark dataset commonly used for 3D BEV Object Detection in autonomous driving.}
\begin{center}
\begin{tabular}{|l|c|c|c|c|c|}
\hline
\textbf{Dataset Senosr Setup} & Kitti \cite{kitti} & BDD \cite{bdd_18} & WOD 
\cite{wod} & nuScenes \cite{nuscenes} & STF \cite{stf} \\
\hline
RGB Cameras & 2 & 1 & 5 & 6 & 2 \\
RGB Resolution & $1242*372$ & $1280*720$ & $1920*1080$ & $1600*900$ & $1920*1024$ \\
Lidar Sensors & 1 & $\times$ & 5 & 1 & 2 \\
Lidar Resolution & 64 & 0 & 64 & 32 & 64 \\
Radar Sensor & $\times$ & $\times$ & $\times$ & 4 & 1 \\
Frame Rate & 10 Hz & 30 Hz & 10 Hz & 10 Hz & 10 Hz \\
\hline
\end{tabular}
\label{dataset_stats}
\end{center}
\end{table*}

\subsection{Evaluation Metrics}
3D object detectors use multiple criteria to measure performance of the detectors viz., precision and recall. However, mean Average Precision (mAP) is the most common evaluation metric. Intersection over Union (IoU) is the ratio of the area of overlap and area of the union between the predicted box and ground-truth box. An IoU threshold value (generally 0.5) is used to judge if a prediction box matches with any particular ground-truth box. If IoU is greater than the threshold, then that prediction is treated as a True Positive (TP) else it is a False Positive (FP). A ground-truth object which fails to detect with any prediction box, is treated as a False Negative (FN). Precision is the fraction of relevant instances among the retrieved instances; while recall is the fraction of relevant instances that were retrieved.
\begin{equation}
Precision=TP/(TP+FP)
\end{equation}
\begin{equation}
Recall=TP/(TP+FN)
\end{equation}
Based on the above equations, average precision is computed separately for each class. To compare performance between different detectors (mAP) is used. It is a weighted mean based on the number of ground-truths per class. \\ \\
In addition, there are a few dataset specific metrics viz., KITTI introduces Average Orientation Similarity (AOS), which evaluates the quality of orientation estimation of boxes on the ground plane. mAP metric only considers 3D position of the objects, however, ignores the effects of both dimension and orientation. In relation to that, nuScenes introduces TP metrics viz., Average Translation Error (ATE), Average Scale Error (ASE) and Average Orientation Error (AOE). WOD introduces Average Precision weighted by heading (APH) as its main metric. It takes heading/ orientation information into the account as well. Also, given depth confusion for 2D-sensors like camera, WOD introduces Longitudinal Error Tolerant 3D Average Precision(LET-3D-AP), which emphasizes more on lateral errors than longitudinal errors in predictions.

\section{Input-data Formats}
\label{input_data_section}
In this section we will cover raw data format returned by camera and radars, and meta data used to get them into the unified coordinate system i.e. egocentric Cartesian coordinate system. 

\subsection{Cameras}
Surround-view camera images can be represented by ${\mathbf{I} \in \mathbb{R}\textsuperscript{N×V×H×W×3}}$. Here, N,V, H and W are the number of temporal-frames, number views, height and width respectively. \\
Given \emph{V} camera images  ${\mathbf{X_{k}} \in \mathbb{R}\textsuperscript{3xHxW}}_V$, each with an extrinsics matrix ${\mathbf{E_k} \in \mathbb{R}\textsuperscript{3x4}}$ and an intrinsics matrix ${\mathbf{I_k} \in \mathbb{R}\textsuperscript{3x3}}$, we can find a rasterized BEV map of the feature in BEV coordinate frame as ${\mathbf{y} \in \mathbb{R}\textsuperscript{CxXxY}}$, where C, X, and Y are channel depth, and height and width of BEV map. The extrinsic and intrinsic matrices together define the mapping from reference coordinates $(x, y, z)$ to local pixel coordinates $(h, w, d)$ for each of the $V$ camera views. Refer to Fig. \ref{nuscenes_data} for surround image view on an autonomous car. 

\subsection{RADARs}
\begin{figure}[ht]
\vskip 0.2in
\begin{center}
\centerline{\includegraphics[width=\columnwidth]{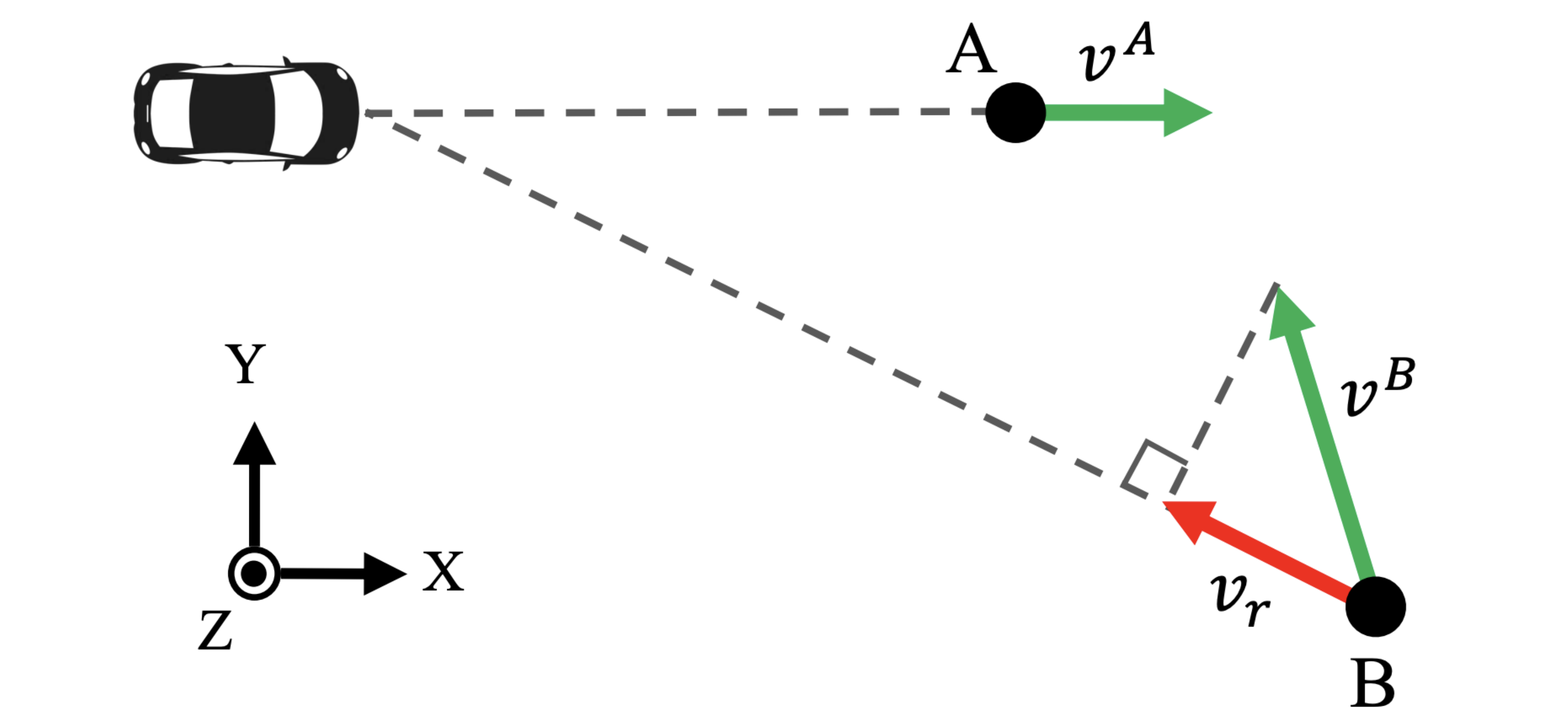}}
\caption{\cite{centerfusion} shows difference between actual and radial velocity. For target A, velocity in the vehicle coordinate system and the radial velocity are the same $(v_A)$. For target B on the other hand, radial velocity $(v_r)$ as reported by the radar is different from the actual velocity of the object $(v_B)$ in the vehicle coordinate system}
\label{radial_vel_fig}
\end{center}
\vskip -0.2in
\end{figure}

Radars are another set of active sensors used in robotics which transmit radio waves to sense the environment and measure the reflected waves to determine the location and velocity of objects. Raw output from the sensor is in polar coordinates, which can be easily converted to BEV space with sensor calibration matrices. However, noisy radar points have to go through filtering which would utilize some form of clustering and temporal tracking. This temporal tracking can be achieved by Kalman filters \cite{kalman}. Kalman filters is a recursive algorithm, which can estimate the current state of the target by obtaining the previously observed target state estimation and the measured value of the current state. After running internal filtering they return 2D points in BEV (without height dimension), providing azimuth angle and radial distance to the object. It also produces radial velocity vector component per 2D point, as shown by \cite{centerfusion} in Fig. \ref{radial_vel_fig}. Here points can be treated as detected objects.
In modern BEV sensor fusion research work, radar detections are represented as a 3D point in the egocentric coordinate system. This 3D point in the radar point-cloud is parameterized as $P = (x, y, z, v_x, v_y)$ where $(x, y, z)$ is the position and $(v_x, v_y)$ is the radial velocity of the object in the $x$ and $y$ direction. This radial velocity is a relative velocity, hence it needs to be compensated with ego vehicle's motion. Due to the high sparsity of this radar point-cloud, we generally aggregate 3-5 temporal sweeps. It adds a temporal dimension to the point cloud representation. Since in lot of approaches detection head runs on the $360^{\circ}$ surround-scene, we merge 3D points from all the radars around the vehicle into a single merged point-cloud.
The nuScenes \cite{nuscenes} dataset provides the calibration parameters needed for mapping the radar point clouds from the radar coordinate system to the egocentric coordinate frame. Refer to Fig. \ref{nuscenes_data} for radar point-cloud from an autonomous car.

\section{Camera-RADAR Fusion}
\label{cam_radar_section}
\begin{figure}[ht]
\vskip 0.2in
\begin{center}
\centerline{\includegraphics[width=\columnwidth]{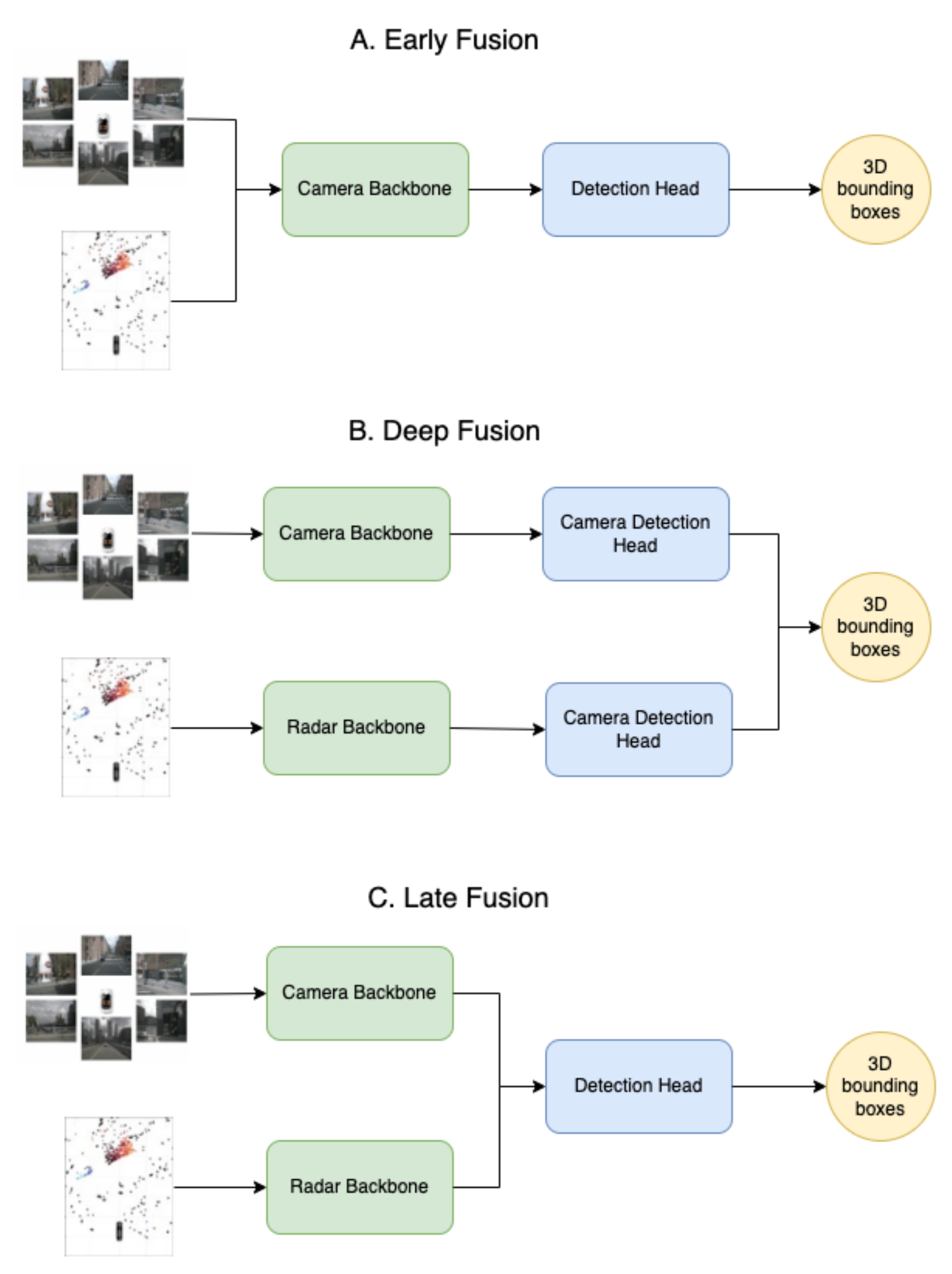}}
\caption{Modality Fusion Methods viz., A. Early Fusion; B. Deep Fusion; C. Late Fusion.}
\label{fusion_methods}
\end{center}
\vskip -0.2in
\end{figure}

Based on at what stage we fuse information of two sensors, these methods can be categorized into three classes viz., early, late and deep fusion. The early and late fusions both have only one interactive operation of different features, which is processed either at the beginning or at the end of the module. However, the deep fusion has more interactive operations of different features. These three approaches can be easily summarized in Fig. \ref{fusion_methods}.

\subsection{Early Fusion}
Early fusion is also referred as data-level fusion. It is the least explored option out of the three. In this approach information from both the sensors is fused together at very early stage i.e. before computation of any features. One of the key challenges in this approach is the synchronization of data. We have cameras and radars data which comes in different coordinate space and moreover nature of the data is also quite orthogonal, where former is a densely packed 2D pixels and later is sparsely packed BEV point cloud. This approach has minimal data-loss issues however, there is no effective way to handle the complexity of aggregating raw data from camera and radar. 
The common line of work in this fusion category is generally done sequentially. Here we first extract region of interests (ROI) based on radar point; then project them on camera; and use some heuristics to gather camera features in the region as done in \cite{early_1}, \cite{early_2}. This is not very reliable approach as there is a high probability that critical objects might get filtered out in radar point-cloud beforehand and due to the nature of the design we won't even look for those objects in the images. However, added benefit with this approach is that we will only run convolution operations on the part of the image that lies inside ROI, hence saving us some compute budget. 

\subsection{Late Fusion}
Late fusion stream of work is the easiest of the three, which makes it the most common approach since past decade of work on camera-radar fusion based object detection. It is pretty certain from our previous reasoning that some of the objects and attributes are better handled by cameras and others by radars. This method lets lets respective sensors detect objects, that they does best and fuse 2 sets of detections together as 1 set of detections using trivial data association techniques \cite{gautam2021sdvtracker}. However, this approach is not able to leverage the fact that features in one sensor-detector won't be able to be augmented by features of the other detector. For example cameras in general can detect boundaries very well and radars can detect velocity with good confidence. Work in this stream can be further classified in two section:\\
\subsubsection{Probabilistic Reasoning Based}
In this approach Bayesian tracking method tracks multi-agent targets with probability-density for multi-modes. It approximates each mode with component probability density. Bayesian algorithm and Particle filter (PF) handles the non-linearities and non-gaussian estimations. It is an iterative algorithm which recursively estimates the state of multiple targets and determine the current target number using maximum-likelihood. Refer \cite{late_1} and \cite{late_2} for representative work. \\
\subsubsection{Kalman Filter Based}. 
In this approach we estimate the current state of the target by obtaining the previously observed target state estimation and the measured value of the current state as in \cite{kalman}. Simple kalman filter can not incorporate nonlinear systems accurately. However, EKF (Extended Kalman Filter) and UKF (Unscented Kalman Filter) are more sophisticated systems that can incorporate non-linearities in the system. EKF linearizes the nonlinear problem, whereas UKF adopts statistical linearization technique to linearize nonlinear function of random variables by sampled points. \\ SORT \cite{sort} and Deep-SORT \cite{deepsort} are seminal papers in this category. SORT explores multi-object tracking task with hungarian matching for data association and constant velocity motion model with kalman fitler estimation. Deep SORT is further extension to this work where authors also add in appearance information in the form of image features in the algorithm. Both these algorithms are very cheap and can be easily handled by the edge-device. MHT \cite{mht} is another tracking-by-detection approach which maintains small list of potential hypotheses, which can be facilitated with the accurate object detectors that are currently available.  \\ \\
Late fusion methods can have the benefit of exploiting the off-the-shelf detection algorithms that are independently developed as modular components. However, late fusion strategies that rely on heuristics and post-processing techniques suffer from performance-reliability trade-offs, especially when these two sensors disagrees. 

\begin{table*}[t]
\caption{Benchmark result for 3D object detection using Camera-radar fusion methods. Abbreviations defined in section \ref{experiments_section}}
\begin{center}
\begin{tabular}{|l|c|c|c|c|c|c|c|c|c|}
\hline
\textbf{Method} & Year & mAP & mATE & mASE & mAOE & mAVE & mAAE & NDS \\
\hline
CenterFusion \cite{centerfusion} & 2020 & 0.326 & 0.631 & 0.261 & 0.516 & 0.614 & 0.115 & 0.449 \\  
CRAFT \cite{craft} & 2022 & 0.411 & 0.467 & 0.268 & 0.456 & 0.519 & 0.114 & 0.523 \\
DETR3d RV & 2022 & 0.439 & 0.537 & 0.264 & 0.408 & 0.477 & 0.119 & 0.539 \\
RCFormer & 2022 & 0.485 & 0.549 & 0.248 & 0.360 & 0.320 & 0.116 & 0.583 \\
\hline
\end{tabular}
\label{experiment_table}
\end{center}
\end{table*}

\subsection{Deep Fusion}
Deep Fusion is also referred as feature-level fusion. In this approach we fuse information of the two sensors  in the form of features, so take it as an intermediate of the previously discussed methods. This approach seems most future promising based on the current research work. This is a learning-based approach, where features from cameras and radars can be computed in-parallel and then soft-associated with each other. This approach can be further classified in three sections: \\
\subsubsection{Radar Image Generation based}
In order to bring radar information into the image domain, radar's features are extracted and transformed into image-like matrix information. This is referred as \emph{radar imagery}. Channels of this radar imagery represents information from the point representation of the radar i.e. physical quantity like distance, speed and so on. \cite{radar_imagery_1}, \cite{radar_imagery_2}, \cite{radar_imagery_3}, \cite{radar_imagery_4} follow this line of work. This approach hasn't been very successful because of inherent sparsity in radar point-cloud which makes them incompatible to formulate into a good image-like matrix. \\
\subsubsection{CNN based}
This line of work focuses on convolution neural networks (CNNs) for doing feature fusion from two different modalities. CNN based detectors used to be SOTA until 2 years ago, until transformer started making contribution on spatial context. In CNN's segment, one of the representative work \cite{cnn_1} uses a neural network that builds on RentinaNet \cite{retinanet} with VGG backbone \cite{vgg}. It uses radar channels to augment the image. This model makes problem simpler by estimating 2D boxes. As authors of \cite{cnn_1} claims that the amount of information encoded in one radar point is different from that of one pixel, we can not just simply early fuse this distinct information. A more optimal solution would be to do it in CNN's deeper layer where information is more compressed and contains more relevant information in the latent space. Since it is hard to abstract that what depth is the right depth for fusion, authors designed a network in a way that it learns itself this fusion strategy. These authors also introduced a technique called \emph{BlackIn} \cite{blackin} where they use dropout strategy but at sensor level instead of neuron level. This helps in leveraging sparse radar points information more which could have been easily shadowed by the dense camera pixels. 

CenterFusion \cite{centerfusion} is another modern work which builds on center-point detection framework \cite{centernet} to detect objects. They solve key data association problem using a novel frustum-based method to associate the radar detections to their corresponding object center. Associated radar detections are used to generate radar-based feature maps to complement the image features and regress to object properties such as depth, rotation and velocity. They claim that just adding the radar input can significantly improve velocity estimation without the need of complex temporal information. Major issue with this work is that it treats primary sensor as camera and will be straight-away discarding detections which are only sensed by the radars. Another problem that we see with this approach is it samples radar points based on BEV center in image. However, there is no guarantee that image network would be able to predict good BEV center due to it's 2D perspective view input-data. \\

\subsubsection{Transformer based}
This line of work typically utilizes transformers module viz., cross-attention to cross-attend features from different modalities and form a finer feature representation. A representative work in CRAFT \cite{craft} associates image proposals with radar point in the polar coordinate system to efficiently handle the discrepancy between the coordinate system and spatial properties. Then in second stage, they use consecutive cross-attention based feature fusion layers to share spatio-contextual information between camera and radar. This paper is one of the SOTA methods on the leader-board \cite{nuscenes} as of date. MT-DETR is another approach which utilizes similar cross-attention structure to fuse cross-modality features.  

\section{Experiments}
\label{experiments_section}
nuScenes \cite{nuscenes} is the widely used datasets in the literature for which sensor setup in Fig. \ref{nuscenes-setup} includes $6$ calibrated cameras and $5$ radars covering the entire $360^{\circ}$ scene. Results on discussed pioneer works are shown on the test set of nuScenes in Table \ref{experiment_table}. This is under the filter \emph{camera-radar track detections}. The key for the metric abbreviations is as follows: mAP: mean Average Precision; mATE: mean Average Translation Error; mASE: mean Average Scale Error; mAOE: mean Average Orientation Error; mAVE: mean Average Velocity Error; mAAE: mean Average Attribute Error; NDS: nuScenes detection score.

\section{Further Extensions}
\label{further_extenstion_section}
Based on the most-recent developments around the production multi-domain BEV perception detections, we will highlight possible directions for the future research.
\subsection{Transformer Extensions}
Looking at the trend in the benchmark datasets, it is pretty apparent that transformer based networks are able to establish right modelling between vision and radar data for getting good fused-feature representation. Even in vision-only based approaches transformers are ahead of there convolutional counterparts. As highlighted in \ref{experiment_table} DETR3D \cite{detr3d} and BEVFormer \cite{bevformer} can be easily extended to initiate queries from radar point-cloud as well. Instead of cross-attending to just vision features, a new cross-attention layer can be added for radar imagery. 

\subsection{Collaborative Perception}
A relatively new field of area is how to make use of multi-agents, mutli-modal transformers to enable collaborative perception. This setup requires a minimal infrastructure setup to enable smooth communications between different autonomous vehicle on the road. CoBEVT \cite{cobevt} shows initial proof of how Vehicle-to-Vehicle communication may lead to superior perception performance. They test their performance on OPV2V \cite{op2v} benchmark dataset for V2V perception.

\section{Conclusion}
\label{conclusion_section}
For the autonomous vehicle's perception reliability, 3D object detection is one of the key challenge which we need to solve. In-fact, to make this problem even harder, we need to do this with sensors which are affordable enough to extend this technology to masses, thereby proving that the life-time cost of an AV is less than that of a driver-operated cab/ vehicle. Camera and RADAR are one of the key sensors that we can leverage to achieve this target. In this paper, we first covered background information to understand why it makes good technical as well as business sense to use cameras and radars for BEV object detection. Then we went into more details on how camera and radar input data is represented. Then we cover state-of-the-art techniques used in the literature and industry for camera-radar fusion by sub-grouping them so that readers can easily follow through. We hope our work will inspire future research on camera-radar fusion for 3D object detection.


\bibliographystyle{ieeetr}
\bibliography{ref}

\end{document}